\pdfoutput=1
\documentclass[12pt]{l4dc2020} 
\title{Learning to Correspond Dynamical Systems}
\usepackage{times}
\usepackage{float}
\usepackage{caption}
\usepackage{subcaption}
\usepackage{amssymb}
\usepackage{etoolbox}

\newbool{@photoready}
\setbool{@photoready}{true}
\newcommand{\zhaoming}[1]{\ifbool{@photoready}{}{\textcolor{green}{Zhaoming: #1}}}
\newcommand{\namhee}[1]{\ifbool{@photoready}{}{\textcolor{red}{Nam Hee: #1}}}
\newcommand{\daniele}[1]{\ifbool{@photoready}{}{\textcolor{blue}{Daniele: #1}}}

\coltauthor{\Name{Nam Hee Kim} \Email{nhgk@cs.ubc.ca}
 \AND
 \Name{Zhaoming Xie} \Email{zxie47@cs.ubc.ca}
 \AND
 \Name{Michiel {van de Panne}} \Email{van@cs.ubc.ca}\\
\addr{Department of Computer Science, The University of British Columbia}
}
\begin{document}
\maketitle
\vspace{-.5in}
\begin{abstract}
 Many dynamical systems exhibit similar structure, as often captured
 by hand-designed simplified models that can be used for analysis and control.  
 We develop a method for learning to correspond pairs of dynamical systems via a learned latent dynamical system. 
 Given trajectory data from two dynamical systems, we learn a shared latent state space and a shared latent dynamics model, 
 along with an encoder-decoder pair for each of the original systems. 
 With the learned correspondences in place, we can use a simulation of one system to produce an imagined motion of its counterpart.  
 We can also simulate in the learned latent dynamics and synthesize the motions of both corresponding systems, as a form of bisimulation.
We demonstrate the approach using pairs of controlled bipedal walkers, as well as by pairing a walker with a controlled pendulum.
 

\end{abstract}
\section{Introduction}
\begin{figure}[b]
    \centering
    \includegraphics[height=2.5cm]{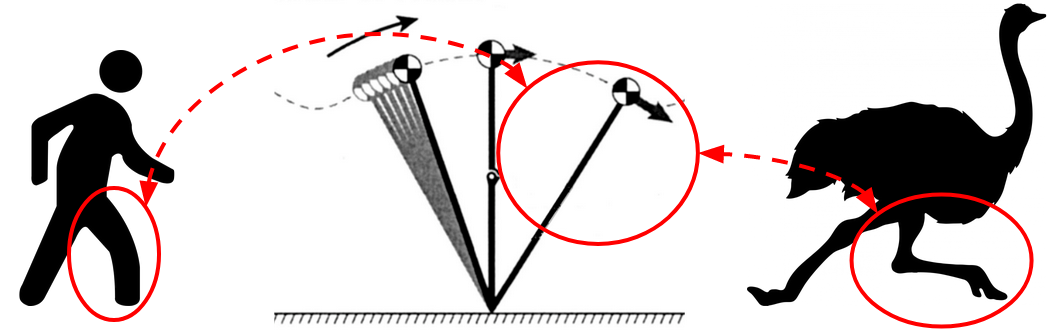}
    \caption{Similar dynamical systems, such as a walking human and an ostrich (left, right) can be mapped to a canonical system such as an inverted pendulum (center). These are often developed by hand. In this paper we show how to learn an abstract shared latent dynamics model.
    }
    \label{fig:human_ostrich_motivation}
\end{figure}
Naturally occurring movement patterns such as locomotion are inherently similar across different species \citep{alexander1983dynamic}. For example, although a human and an ostrich have very different anatomical features, both share a fundamental system of joints that work together to produce bipedal gaits, and possibly share similar foot-placement strategies for balance recovery in the face of a large disturbance. In principle, knowing the similarities across these bipedal locomotion patterns should be beneficial for building a controller for a new bipedal robot, i.e., we could use a known control strategy on one system to control another. This is
also a motivation for using simplified models, such as an inverted pendulum model, as they provide crude-but-canonical models of the dynamics that can be used to guide the control. While finding good correspondences is a well-established problem in other settings,
including matching geometric shapes, there has been much less work examining how to learn correspondences across multiple dynamical systems.


In this paper, we introduce a simple supervised learning method named L2CDS (Learning to Correspond\footnote{although ``correspond'' is an intransitive verb, we use the phrase ``to correspond'' as a short-hand way of saying ``putting into correspondence''.} Dynamical Systems). 
L2CDS is a data-driven approach that can put two-or-more dynamical systems into correspondence by extracting a shared 
latent state space that encapsulates the similarities across the motions of the systems via a learned latent state-space that
is shared across systems, together with a learned latent dynamics model. 
We can project the state of a source system to a desired target system 
by encoding to the corresponding latent space and then decoding to the corresponding target state space.

Our contributions are as follows. We formalize the problem of dynamical system correspondence and introduce a learned latent space dynamics model to solve it in a modular fashion. The autoencoder-based model finds dynamics-compatible correspondences across the states of multiple dynamical systems. Finally, we give results for well-studied controlled dynamical systems and evaluate their correspondence both qualitatively and quantitatively.
We refer the reader to the project website: \url{https://sites.google.com/view/l2cds}
for a narrated video that provides explanations of the method and animated results.

\section{Related Work}
\paragraph{Correspondence} We adopt the view of correspondence as a matching between different modalities of a common latent object. Iterative closest point (ICP) \citep{besl1992method} is a classical method to solve shape correspondence. 
There is also significant work on motion retargeting, where kinematic sequences are transferred between characters with different morphologies, e.g. \citep{Jain2011MotionA} \citep{Rhodin-2014-Motion_Mapping}.
These do not deal with the general problem of learning to correspond dynamical systems, which is the focus of our work.

\paragraph{Template models for locomotion} Due to the high number of degrees of freedom of typical bipeds, 
simplified models are often used to model and control these dynamical systems (See Figure~\ref{fig:human_ostrich_motivation}). 
This includes inverted pendulums and their many variants, e.g. \citep{LIP-2013, capture_point-2006,SLIP-2006}. 
These models are utilized to design simple control laws for the original complicated bipedal systems. 
Robots having different morphologies can also be put into correspondence to simplify control. 
For example, \citet{Raibert_1986} use correspondences between one-legged hopper gaits, biped gaits, and quadruped gaits to 
produce locomotion policies that share significant structure. The aforementioned methods involve hand-crafting the reduction from a complex system to a simpler one with lower degrees of freedom. Instead, our work focuses on leveraging state trajectory data 
to learn an appropriate reduction.

\paragraph{Autoencoders with latent dynamics} Learning the dynamics of the motion and the physical environment is a classic learning problem in the control literature. Early arguments for learned dynamics can be traced to the 1980s, e.g. \citep{werbos1987learning}. \citet{watter2015embed} instead learned the dynamics on the latent space produced via variational autoencoders and demonstrated that the latent dynamics can be utilized for easier control. \citet{ha2018world} proposed using learned latent dynamics to augment reinforcement learning. More recently, \citet{hafner2019dream} used the latent system exclusively to learn the control policy of the original system. Our work concerns using latent dynamics as a mechanism to achieve correspondence with temporal consistency, to aid the correspondence of dynamical systems. A future goal is to leverage the learned latent dynamics to improve the efficiency and stability of reinforcement learning in continuous control.

\paragraph{Learning latent features for optimal control} Our work is similar in spirit to \citet{gupta2017learning}, which used proxy tasks and a similarity loss to induce an overlap in the latent feature space across two characters. The authors further used the learned latent space to improve reinforcement learning performance. While \citet{gupta2017learning} focuses on the transfer of new skills between a fixed number of agents, we focus on the possibility of the transfer of a fixed number of skills from expert agents to a novice agent. Moreover, we improve upon the strong prior knowledge used to establish the correspondences in~\citet{gupta2017learning} by learning (possibly non-unique) correspondences explicitly with the use of latent dynamics.

\section{Method}

\begin{figure}
\centering
\includegraphics[width=\textwidth]{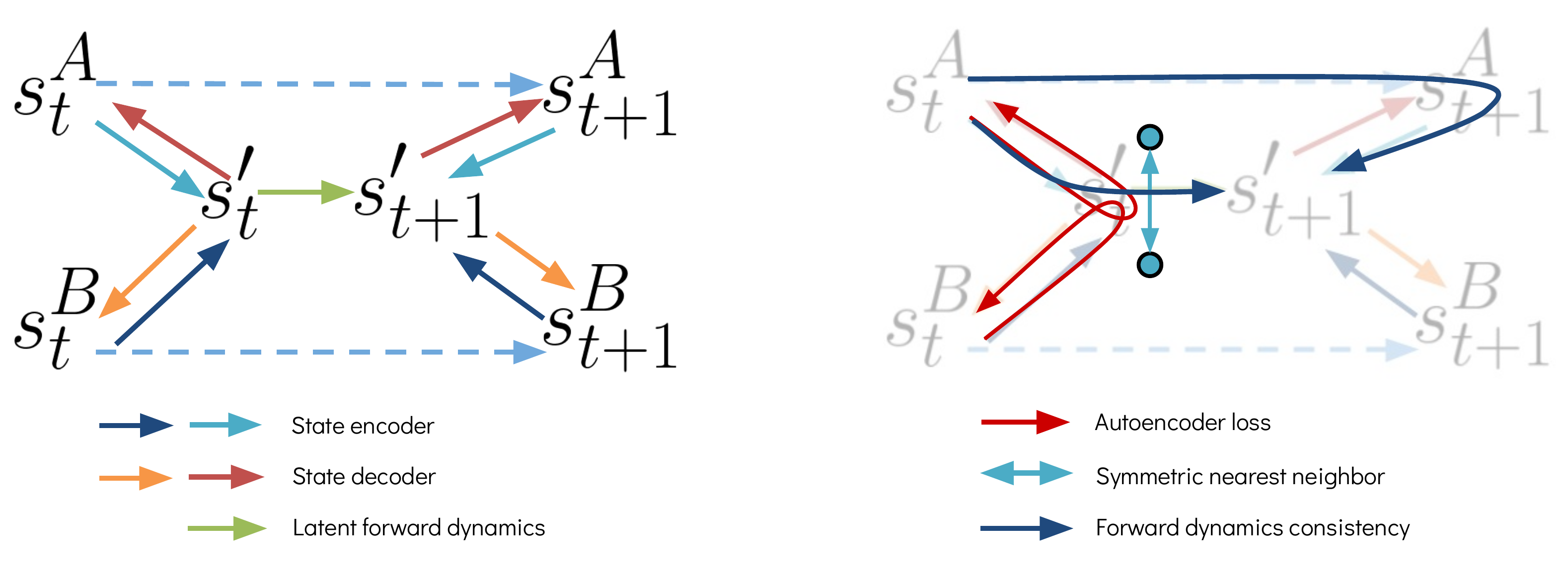}
\caption{Left: Summary figure for variables and mappings in L2CDS, where each solid arrow represents a learned mapping. Dashed arrows represent the forward dynamics of original state trajectories. Right: summary figure of losses involved in optimizing L2CDS. Traversing the arrows in the left figure results in backpropagation-enabled transformations. Best viewed in color.}
\label{fig:overview}
\end{figure}
\subsection{Problem Formulation}
Consider a discrete-time dynamical system consisting of a tuple $(\mathcal{S}, f)$, with the state space $\mathcal{S} \subseteq  \mathbb R^n$ and the transition function $f: \mathcal{S} \to \mathcal{S}$. The state of the dynamical system $s_{t+1}$ at time $t+1$ can be computed from the state $s_t$ at time $t$ via $s_{t+1} = f(s_t)$. 

Given two dynamical systems $A = (\mathcal{S}^A, f^A)$ and $B = (\mathcal{S}^B, f^B)$, our goal is to find a pair of correspondence functions $C_A^B: \mathcal{S}^A \to \mathcal{S}^B$ and $C_B^A: \mathcal{S}^B \to \mathcal{S}^A$. $C_A^B$ projects a state of the source system $A$ to the state space of the target system $B$. $C_B^A$ is the inverse of $C_A^B$. The correspondence functions must not only map the distributions of states, but also honor the temporal ordering of states based on each system's dynamics. In other words, consecutive states in one system must correspond to consecutive states in the other:
\begin{align}
C_A^B(s_{t+1}^A) &= f^B(C_A^B(s_t^A)), & C_B^A(s_{t+1}^B) &= f^A(C_B^A(s_t^B)).\label{eq:dynamics_consistency}
\end{align}
Dynamical systems can often be embedded in a lower dimensional manifold. Instead of directly learning the corresponding function $C_A^B$, we embed the two dynamical systems into a latent dynamical system $D' = (\mathcal{S}', f')$.  More formally, we first learn the correspondence function $C_A^{D'}$ and $C_{D'}^{B}$ and then compose the two functions $C_{D'}^B \circ C_A^{D'}$ to compute $C_A^B$. See Figure \ref{fig:overview} for a graphical representation of this approach.

\subsection{Collecting State Trajectories}

To learn the correspondence functions in a data-driven fashion, we rely on the collection of state trajectories for each system of interest. For systems A and B, we compile datasets $\mathcal{D}^A$ and $\mathcal{D}^B$ by simulating each system forward. We advance each state trajectory to a desired horizon and then restart the system with some randomness to collect a new one. More formally:
\begin{align}
    \mathcal{D} &= \bigcup_{i=1}^R \{\left< s_t, s_{t+1}\right> \ | \ t = 1, 2, \cdots, H\}_i
\end{align}
where $i$ is the state trajectory index, $H$ is the horizon for each trajectory, and $R$ is the number of resets performed during collection. The same data collection strategy is used for systems A and B.

\subsection{Autoencoders}
We employ autoencoders, e.g., ~\citep{kramer1991nonlinear}, to project original states onto the latent state space and back onto the original state space ($s_t \to s'_t \to \hat s'_t)$. To use the terminology of autoencoders, we denote the corresponding function and its inverse to be $E$ (encoder) and $D$ (decoder) respectively. In this work, the encoder and decoder are represented as a feed forward neural network, thus the decoder is only an approximate inverse. One can project a state in the original space onto the latent space and/or project a latent state onto the original space with the following set of operations:
\begin{align}
    s'^A_t &= E^A(s^A_t) & \hat s^A_t &= D^A(s'^A_t) & s'^B_t &= E^B(s^B_t) & \hat s^B_t &= D^B(s'^B_t)
\end{align}
Where $E^A$ and $D^A$ are the encoder and the decoder for the states of system A, and so on. Note that the arguments for decoders $D^A$ and $D^B$ are interchangeable, since latent states $s'^A_t$ and $s'^B_t$ share the same space. Reconstructed states $\hat s^A_t$ and $\hat s^B_t$ are used as estimates of the corresponded states for systems A and B at the tail-end of our pipeline. To ensure that the reconstructed state is consistent with the original state, we use the autoencoder loss (Note that states $s_t^A$ and $s_t^B$ are sampled from datasets $\mathcal{D}^A$ and $\mathcal{D}^B$, not decoders):
\begin{align}
    \mathcal{L}_{AE}(\mathcal{D}) = \mathbb{E}_{s^A_t \sim \mathcal{D^A}, s^B_t \sim \mathcal{D^B}} \left[ ||\hat s^A_t - s^A_t||^2 + ||\hat s^B_t - s^B_t||^2 \right] \label{eq:L_AE}
\end{align}
\subsection{Symmetric Nearest Neighbors}
\begin{figure}
    \centering
    \begin{minipage}[t]{.45\textwidth}
    \centering
    \includegraphics[width=\textwidth]{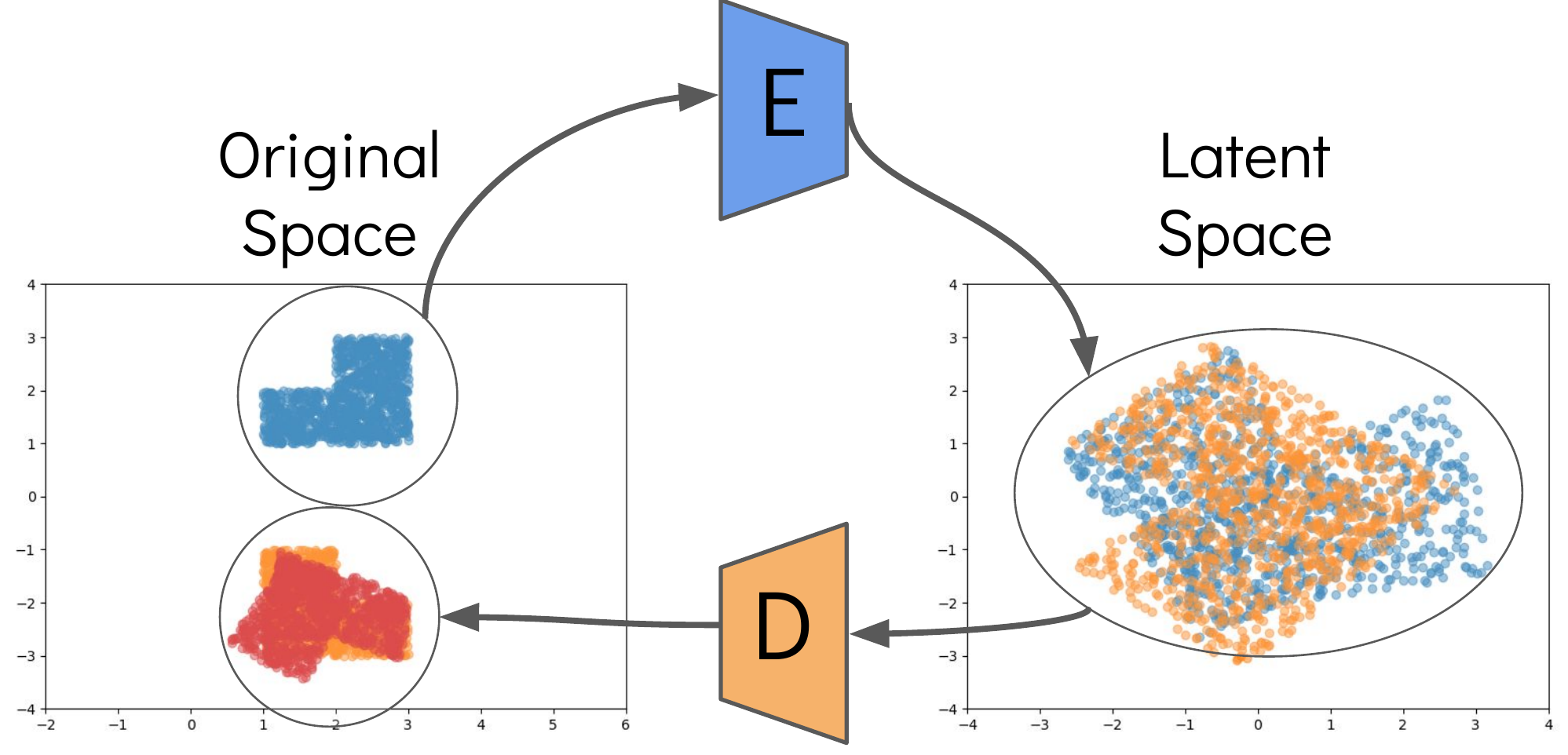}
    \end{minipage}
    \hfill
\begin{minipage}[t]{.45\textwidth}
  \centering
  \includegraphics[width=\textwidth]{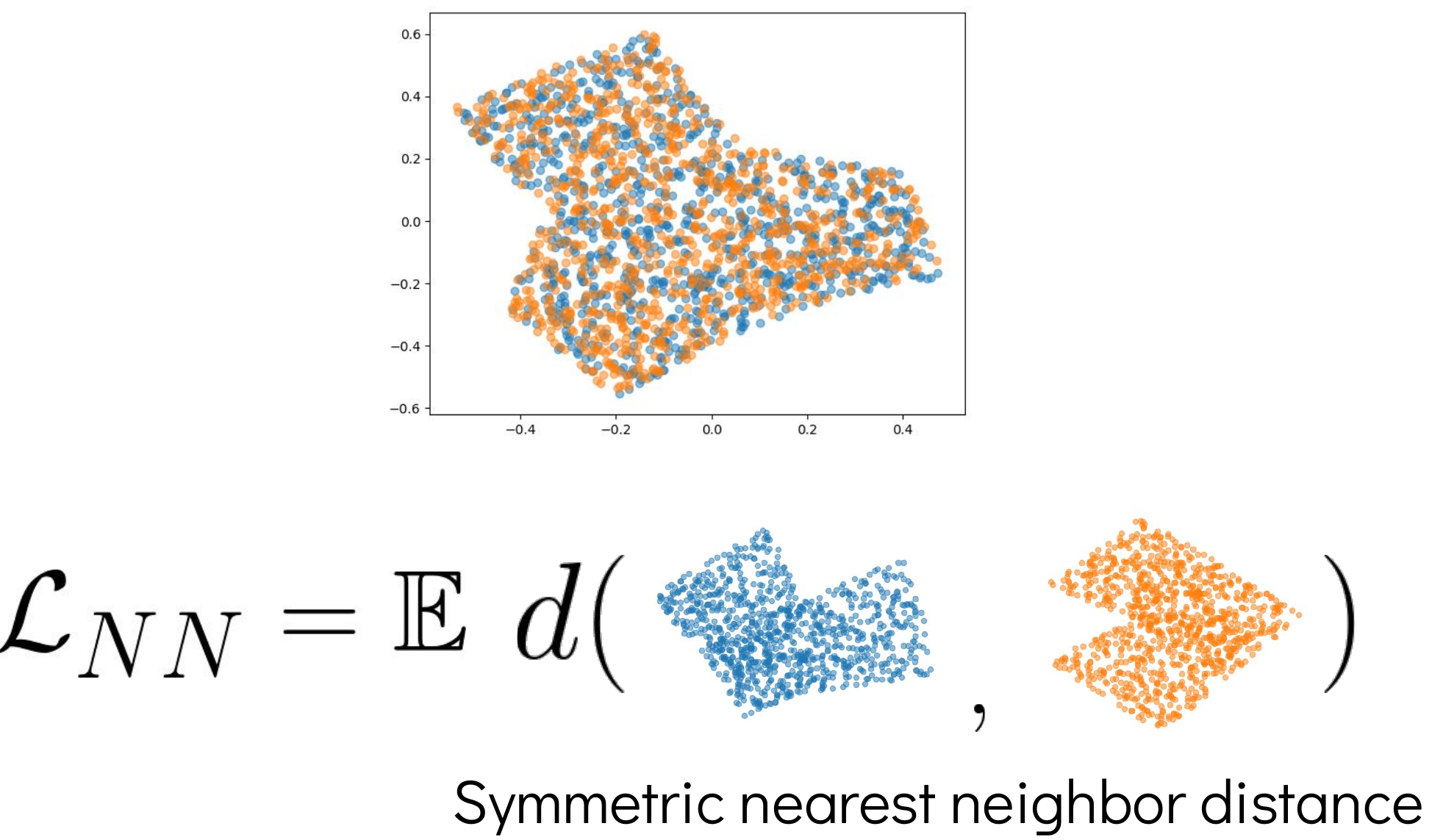}
\end{minipage}
\caption{Left: Na\"ively learned autoencoders do not enforce the latent space to be shared across systems. Right: regularizing for the minimum distance between the two sets of latent space can act to fuse state distributions together. Best viewed in color.}
 \label{fig:symmetric_nn}
\end{figure}
When an encoder-decoder pair is learned for each system, the encoders are not constrained to produce overlapping latent state distributions across the systems of interest. This prohibits using the decoders interchangeably for correspondence 
(see Figure~\ref{fig:symmetric_nn}). We employ a symmetric nearest neighbor loss, where the encoders are penalized for mapping onto subspaces far away from other encoders.



For our two-system example, the nearest-neighbor loss is computed by (1) sampling a separate batch from each dataset, i.e. $s^A_t \sim \mathcal{D}^A$ and $s^B_t \sim \mathcal{D}^B$, (2) computing latent states $s'^A_t$ and $s'^B_t$ for each batch, and (3) computing the minimum distance between each state in system A and another state in system B, and vice versa. We minimize the expected minimum distances by introducing the following loss term:
\begin{align}
    \mathcal{L}_{NN}(\mathcal{D}) = \mathbb{E}_{s^A_t \sim \mathcal{D}^A, s^B_t \sim \mathcal{D}^B}
    \left[
    {\min_{s^B_t} || s'^A_t - s'^B_t ||^2} 
    + 
    {\min_{s^A_t} || s'^A_t - s'^B_t ||^2}
    \right]
\end{align}
The right side of Figure \ref{fig:symmetric_nn} illustrates the effect of this loss function.

\subsection{Latent Forward Dynamics}

Up to this point, we have enforced different systems to project their states onto the same latent state space and required latent states of different sources to reside in close proximity. However, although the previous loss terms enforce spatial consistency in latent state distribution, the resulting temporal ordering of the latent states is highly ambiguous, without the dynamics constraint as required by Equation \eqref{eq:dynamics_consistency}. We prevent this degeneracy by introducing a latent dynamics model that is shared across the systems of interest. The latent dynamics model $F$ takes a latent state and produces a displacement to the latent state at next time step:
\begin{align}
    v'_{t} &= F(s'_t), & s'_{t+1} &= s'_t + v'_t.
\end{align}
We encourage the encoder to project states $s_t$ and $s_{t+1}$ to the latent state space and ensure that the encoder's results are consistent with the latent dynamics. The following loss function expresses this objective:
\begin{align}
    \mathcal{L}_{FD}(\mathcal{D}) &= \mathbb{E}_{\left< s_t, s_{t+1} \right> \sim \mathcal{D}} \left[ || \left(s'_t + F(s'_t)\right) - s'_{t+1} ||^2 \right].
\end{align}
Note that the latent forward dynamics model $F$ is learned in a fully unsupervised fashion, essentially providing regularization for $\mathcal{L}_{FD}$. We still optimize $F$ along with other modules such that its predictions are stable.

Furthermore, a state in a Markovian dynamical system must describe all information necessary for predicting the state at the next timestep. Practically, a state $s_t$ typically breaks down into the pose $\theta_t$ and the velocity $\dot\theta_t$ of the system. Then the dynamics function establishes a differential equation involving $\theta_t$, $\dot\theta_t$, and $\ddot\theta_t$ (acceleration) of the system. We reflect this in the latent space, so that the latent system also honors the Markovian definition of a state. In other words, we encourage the encoders to produce latent states that observe the relationship between pose and velocity with respect to time. More formally, let
\begin{align}
    \begin{bmatrix}
    \theta_t'\\
    \dot \theta_t'
    \end{bmatrix} = s'_t, &\qquad
    \begin{bmatrix}
    \theta_{t+1}'\\
    \dot \theta_{t+1}'
    \end{bmatrix} = s_{t+1}'
\end{align}
where $\theta_t'$ and $\dot \theta_t'$ are the pose and velocity of the latent system at time $t$, respectively. If $s_t' \in \mathbb R^{2k}$, then $\theta_t', \dot\theta_{t}' \in \mathbb R^k$. We minimize the following loss:
\begin{align}
    \mathcal{L}_{PV}(\mathcal{D}) &= \mathbb{E}_{\left< s_t, s_{t+1} \right> \sim \mathcal{D}} \left[ || (\theta_t' + \dot \theta_t') - \theta_{t+1}'||^2 \right]
\end{align}
to encourage the encoders to honor the pose-velocity relationship. 
\subsection{Final Loss Function and Runtime Procedure}

Our final loss function is the weighted sum of the four loss functions introduced above:
\begin{align}
    \mathcal{L}(\mathcal{D}) &= \lambda_{AE} \mathcal{L}_{AE}(\mathcal{D}) + \lambda_{NN} \mathcal{L}_{NN}(\mathcal{D}) + \lambda_{FD} \mathcal{L}_{FD}(\mathcal{D}) + \lambda_{PV} \mathcal{L}_{PV}(\mathcal{D})
\end{align}
where loss weights $\lambda_{AE}$, $\lambda_{NN}$, $\lambda_{PV}$, and $\lambda_{FD}$ are selected experimentally so that the resulting correspondence is qualitatively and quantitatively sound. The right part of Figure \ref{fig:overview} summarizes this final loss function. After optimizing over the final loss function, correspondence is achieved by using the learned mappings to project source system states to the latent space, and then projecting the latent states to the target system state space. In other words, we can put states into correspondence by using the learned decoders interchangeably.
\section{Experiments} \label{experiments}

\begin{figure}[t]
\centering
\begin{subfigure}
    \centering
    \includegraphics[height=2cm]{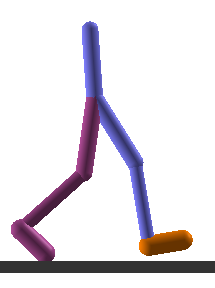}\hspace*{0.9cm}
\end{subfigure}
\begin{subfigure}
    \centering
    \includegraphics[height=2cm]{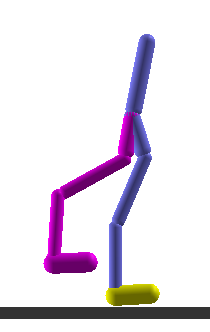}\hspace*{0.9cm}
\end{subfigure}
\begin{subfigure}
    \centering
    \includegraphics[height=2cm]{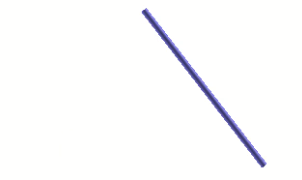}
    \label{fig:pendulum}
\end{subfigure}
\caption{The dynamical systems for our experiments: (a) \texttt{walker2d}: a 2D walker character with 6 joints. $|s_t| = 17$. (b) \texttt{ostrich2d}: a 2D walker character with 8 joints. $|s_t| = 21$. (c) \texttt{pendulum}: a 1-DOF pendulum system driven to a stable cyclic motion using a PD controller.}
\label{fig:characters}
\end{figure}


We demonstrate our results on the three systems shown in Figure~\ref{fig:characters}.  
We put these systems into correspondence using the L2CDS pipeline. These systems are simulated with DART~\citep{lee2018dart}. The neural networks and learning algorithms are implemented in PyTorch \citep{steiner2019pytorch}. RAdam \citep{liu2019variance} is used to optimize the neural network parameters. We use $256 \times 256$ multi-layer perceptrons (MLPs), where ReLU is used as hidden layer activation and Tanh is applied at the output to limit the output range. We use a batch size of 4096 for stochastic gradient descent; note that this is high compared to what is typically used for other learning tasks but is necessary in our settings to accommodate the needed sampling accuracy for the symmetric nearest-neighbor loss. To prevent state decoders and forward dynamics models from learning to separate latent states of different sources, we add Gaussian noise to the inputs of the decoder networks and the latent forward dynamics model. The effect of the noise is significant, as summarized in Table \ref{table:ablation}.

The dynamical systems we consider also involve control inputs. To make the dynamical systems fully autonomous, we pre-train a policy for each system. In particular, the pendulum is running a PD controller that tracks a time-varying PD target that repeats at a fixed period, \texttt{walker2d} and \texttt{ostrich2d} run policies that track repeating walking cycles, which are obtained using DeepMimic \citep{2018-TOG-deepMimic}. When generating the state trajectories, we add small Gaussain noise 
to the policies as to encourage more variety in the states visited. 

Quantitatively, we evaluate the estimated correspondence by measuring the difference between the distributions of the original states $s_t$ and the reconstructed states $\hat s_t$ after using the learned projections in L2CDS $(s_t \to s'_t \to \hat s_t)$. As a heuristic to this difference, we use the mean symmetric nearest neighbor (MSNN) distances within the original state space:
\begin{align}
    \mathcal{E}_{MSNN} = \frac{1}{2n} \left(\sum_{s_t} \min_{\hat s_t} ||s_t - \hat s_t|| + \sum_{\hat s_t} \min_{s_t} ||s_t - \hat s_t||\right) \label{eq:msnn}
\end{align}
where $n$ is the number of states stored in the dataset.

Qualitatively, we observe similarities in corresponding states of different systems. Namely, we simulate one character with the pre-trained policy and use the learned mappings to generate the estimated corresponding states of another system. We particularly focus on the effect of perturbations in the simulated system. A good correspondence across dynamical systems must take into account the system's interactions with its physical environment, such as contact and forces. Therefore, we expect the correspondence to go beyond a simple mapping between multiple limit cycles; rather, the dynamics must capture and put perturbed states of systems into correspondence.

\subsection{Corresponding 2D Walker with Pendulum}

Our method successfully learns to correspond the classical 2D walker character with the pendulum. The supplementary video shows that the swing of the pendulum is aligned with the gait of the walker, such that one swing of the pendulum completes as one walking cycle of the walker is completed. The limit cycle drawn with the estimated corresponding states of the pendulum shows that both the position and the velocity of the pendulum are reconstructed with high precision. Finally, perturbing the walker with an external force results in the perturbation of the pendulum, showing that the state-to-state mapping goes beyond a simple matching of two cycles.

\subsection{Corresponding 2D Walkers}


Table \ref{table:ablation} shows $\mathcal{E}_{\mathrm{MSNN}}$ computed on various configurations on the walker-ostrich correspondence case. We ablate each logical component of the full loss function and also show the impact of the presence of noise in the inputs of state decoders and forward dynamics models. We use Equation \eqref{eq:msnn} on a test dataset of $n=1000$ state tuples, generated with a random seed different from that of the training data (otherwise exact same procedure). 

In Table \ref{table:ablation}, WLW (walker-latent-walker) and OLO (ostrich-latent-ostrich) are the two projections that consistently achieve low $\mathcal{E}_{\mathrm{MSNN}}$. This is expected since state reconstruction in these two projections is directly correlated with the autoencoder loss $\mathcal{L}_{\mathrm{AE}}$ in Equation \ref{eq:L_AE}. Once the inter-system correspondence is involved (i.e. WLOLW, OLWLO, WLO, OLW), $\mathcal{E}_{\mathrm{MSNN}}$ increases as the latent state distributions are no longer identical. Without $\mathcal{L}_{\mathrm{NN}}$, the model can still achieve low $\mathcal{E}_{\mathrm{MSNN}}$ for WLW and OLO,
but making large error for other projections. 

The lowest $\mathcal{E}_{\mathrm{MSNN}}$ involving inter-system correspondence is achieved by learning the model without the latent forward dynamics components. However, visualizing the results, we see that this model yields a qualitatively poor correspondence, due to the ambiguities induced by the lack of temporal consistency in the latent states. On the other hand, adding noise to state decoders and the latent forward dynamics model has a significant effect on the reduction of $\mathcal{E}_{\mathrm{MSNN}}$. This is thanks to state encoders producing latent states that state decoders and forward dynamics models cannot distinguish, encouraging a homogeneous mixture of the latent states of different sources. We note that the full model with noise achieves the next best performance in terms of $\mathcal{E}_{\mathrm{MSNN}}$, and has the most qualitatively sound result. 
We refer the reader to the video for the qualitative results.

\begin{table}[t]
\centering
\resizebox{\textwidth}{!}{%
\begin{tabular}{l|l|l|l|l|l|l|l}
\hline
              &                & WLW  & OLO  & WLOLW & OLWLO & WLO   & OLW   \\
\hline
With noise    & Full model     & $0.89 (\pm0.09)$ & $0.78 (\pm0.07)$ & $1.07 (\pm0.06)$ & $1.09 (\pm0.10)$ & $1.35 (\pm0.17)$ & $1.30 (\pm0.13)$  \\
              & Without $\mathcal{L}_{\mathrm{FD}}, \mathcal{L}_{\mathrm{PV}}$ & $0.33 (\pm0.01)$ & $0.29 (\pm0.00)$ & $0.47 (\pm0.01)$ & $0.41 (\pm0.02)$ & $1.64 (\pm0.15)$ & $1.44 (\pm0.18)$  \\
              & Without $\mathcal{L}_{\mathrm{NN}}$     & $0.72 (\pm0.00)$ & $0.64 (\pm0.02)$ & $39.00 (\pm25.24)$ & $28.78 (\pm18.41)$ & $26.07 (\pm13.65)$ & $25.65 (\pm10.51)$\\
\hline
Without noise & Full model     & $0.56 (\pm0.04)$ & $0.47 (\pm0.04)$ & $1.57 (\pm0.52)$ & $2.75 (\pm0.77)$ & $3.87 (\pm1.14)$ & $3.40 (\pm0.82)$  \\
              & Without $\mathcal{L}_{\mathrm{FD}}, \mathcal{L}_{\mathrm{PV}}$ & $0.30 (\pm0.01)$ & $0.29 (\pm0.01)$ & $0.58 (\pm0.04)$ & $0.52 (\pm0.05)$ & $2.32 (\pm0.16)$ & $1.88 (\pm0.17)$  \\
              & Without $\mathcal{L}_{\mathrm{NN}}$     & $0.54 (\pm0.02)$ & $0.52 (\pm0.11)$ & $28.90 (\pm18.64)$ & $44.52 (\pm20.76)$ & $43.48 (\pm22.20)$ & $46.89 (\pm22.40)$ \\
\hline              
\end{tabular}}
\caption{MSNN distances (Equation \eqref{eq:msnn}) for walker-ostrich correspondence case. Taken across 5 random seeds for each configuration, reporting mean errors with standard deviations. W=Walker, L=Latent, O=Ostrich. WLW refers to using learned mappings to project walker-latent-walker $(s^{\text{walker}}_t \to s'_t \to \hat s^{\text{walker}}_t)$, and so on. ``Noise'' refers to Gaussian noise being added to each decoder and the latent forward dynamics model.}
\label{table:ablation}
\end{table}
\section{Conclusion}
We have presented L2CDS, a data-driven approach to learning correspondences across the states of multiple dynamical systems. 
L2CDS achieves correspondence by viewing the input systems as different manifestations of a common latent system. 
The latent system is modeled by simultaneously learning the latent state and dynamics based on the trajectories from the given dynamical systems.

A limitation of the current method is that our dynamical systems (controlled physics-simulated characters) 
were designed to complete their limit cycles in an equal time interval, and thus the resulting corresponded
state trajectories can align very well. In real systems, however, we cannot expect such alignments due to possible
mismatches in the time-scales of the dynamics, as would readily result from different physical scales, e.g.,
a cat steps at a much higher frequency than an elephant. Another limitation is that our method does not 
currently allow for outlier rejection. This can be problematic even when matching two different instances of the same system
if the samples collected from each system sample do not come from fully-overlapping state-space regions.

In future work, we wish to accelerate the learning of control policy for novel-but-similar systems,
by only learning the correspondence with the latent dynamical system. 
This is a promising direction for achieving efficient and stable transfer learning via a type of learning-by-analogy. 
Inverted pendulum and centroidal dynamics are widely used to design control policies 
for bipedal robots, and we wish to explore if similar results can be achieved via a learned canonical system.

\bibliography{ref}
\newpage
\artappendix
\section{Is the Learned Latent System Meaningful?}

Enforcing the dynamics in a shared latent space effectively produces a dynamical system that operates on the latent states. Since we aim to leverage this latent system in future work, it is important to assess whether this learned latent system is actually encapsulating something meaningful in learning to correspond dynamical systems. 

\begin{figure}[!ht]
    \centering
    \includegraphics[width=\textwidth]{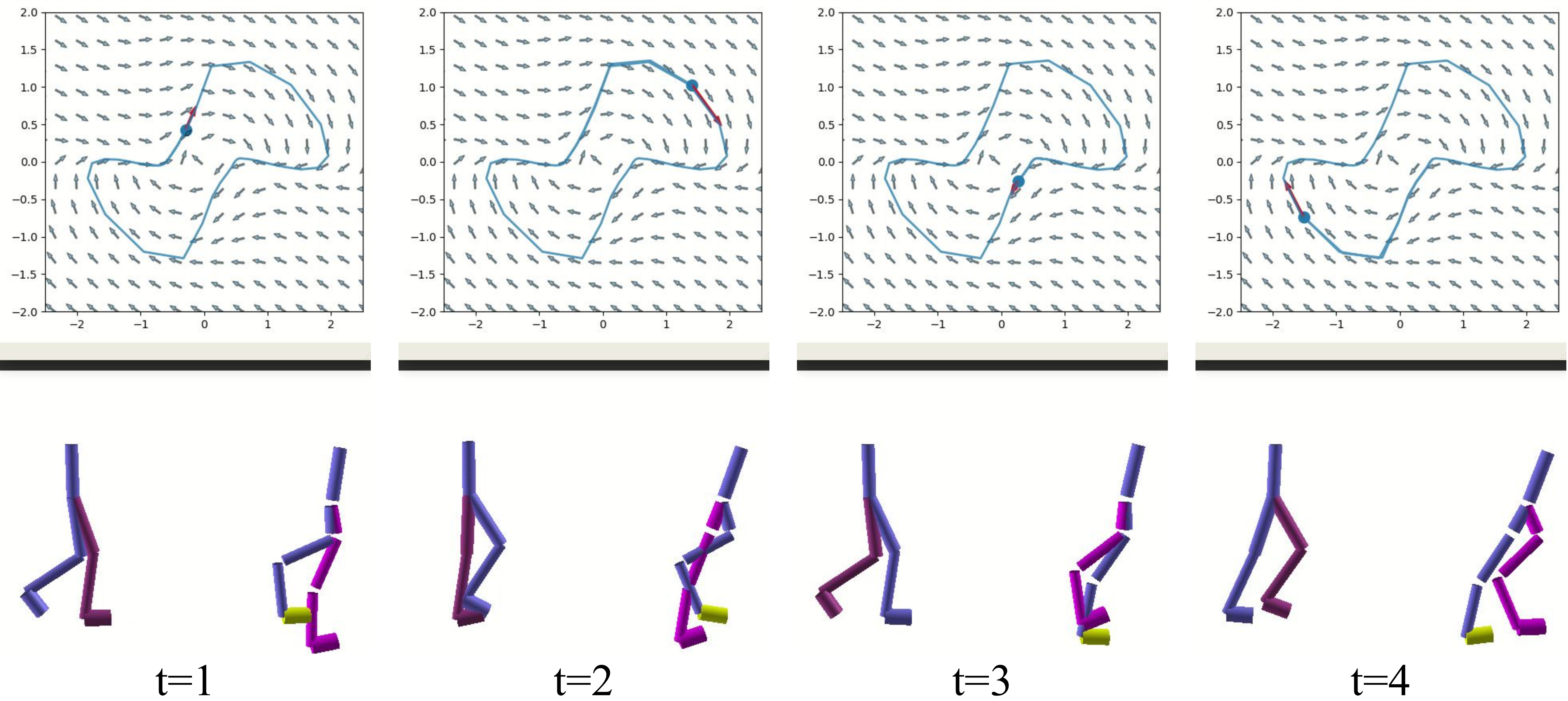}
    \caption{The evolution of the latent system forward in time. Top: the latent system is simulated forward with the latent dynamics as per Equation \ref{eq:latent_sim}. The latent system is learned from applying L2CDS to \texttt{walker2d} and \texttt{ostrich2d}. The latent state trajectory (blue line segments) is visualized in PCA subspace with the first two principal components. The red arrow indicates the output of the latent forward dynamics model encoded into the PCA subspace. The latent forward dynamics is visualized as a vector field in the PCA subspace. Bottom: decoded states are rendered into respective characters. The decoded motions are visually very close to the walking motions of the original systems. See supplmentary video for the animated result.}
    \label{fig:latent_system_meaningful}
\end{figure}

One way to test the validity of this latent system is to check if our view of correspondence corroborates with the learned models. In our work, correspondence is a matching between two modalities of a common latent object. Then the correspondence between two dynamical systems is a matching between two modalities of a common latent dynamical system. Therefore, two corresponding state trajectories of each real system can be viewed as a ''rendered`` version of a latent state trajectory simulated forward by the latent dynamics. We show that the learned dynamics $F$ and the decoders $D^A$ and $D^B$ in L2CDS produce this result in fact. More formally, we choose a starting latent state $s'_1$ and simulate the latent system forward for $T$ timesteps using $F$:
\begin{align*}
s'_{t+1} = s'_t + F(s'_t), \qquad t=1, 2, \cdots, T-1. \label{eq:latent_sim}
\end{align*}
The resulting simulated latent state trajectory $(s'_1, s'_2, \cdots, s'_T)$ can be decoded into each system: 
\begin{align}
    s^A_t &= D^A(s'_t), \qquad t=1, 2, \cdots, T\\
    s^B_t &= D^B(s'_t), \qquad t=1, 2, \cdots, T
\end{align}
We obtain the sequence $(s^A_1, s^A_2, \cdots, s^A_T)$ and $(s^B_1, s^B_2, \cdots, s^B_T)$ accordingly. Rendering the decoded sequences results in a qualitatively sound motion in the real systems. Figure \ref{fig:latent_system_meaningful} shows the latent system's simulated states being decoded into \texttt{walker2d} and \texttt{ostrich2d}.

\section{Are the Perturbed State Mappings Meaningful?}
\begin{figure}[!ht]
\begin{minipage}[t]{.5\textwidth}
    \centering
    \includegraphics[width=\textwidth]{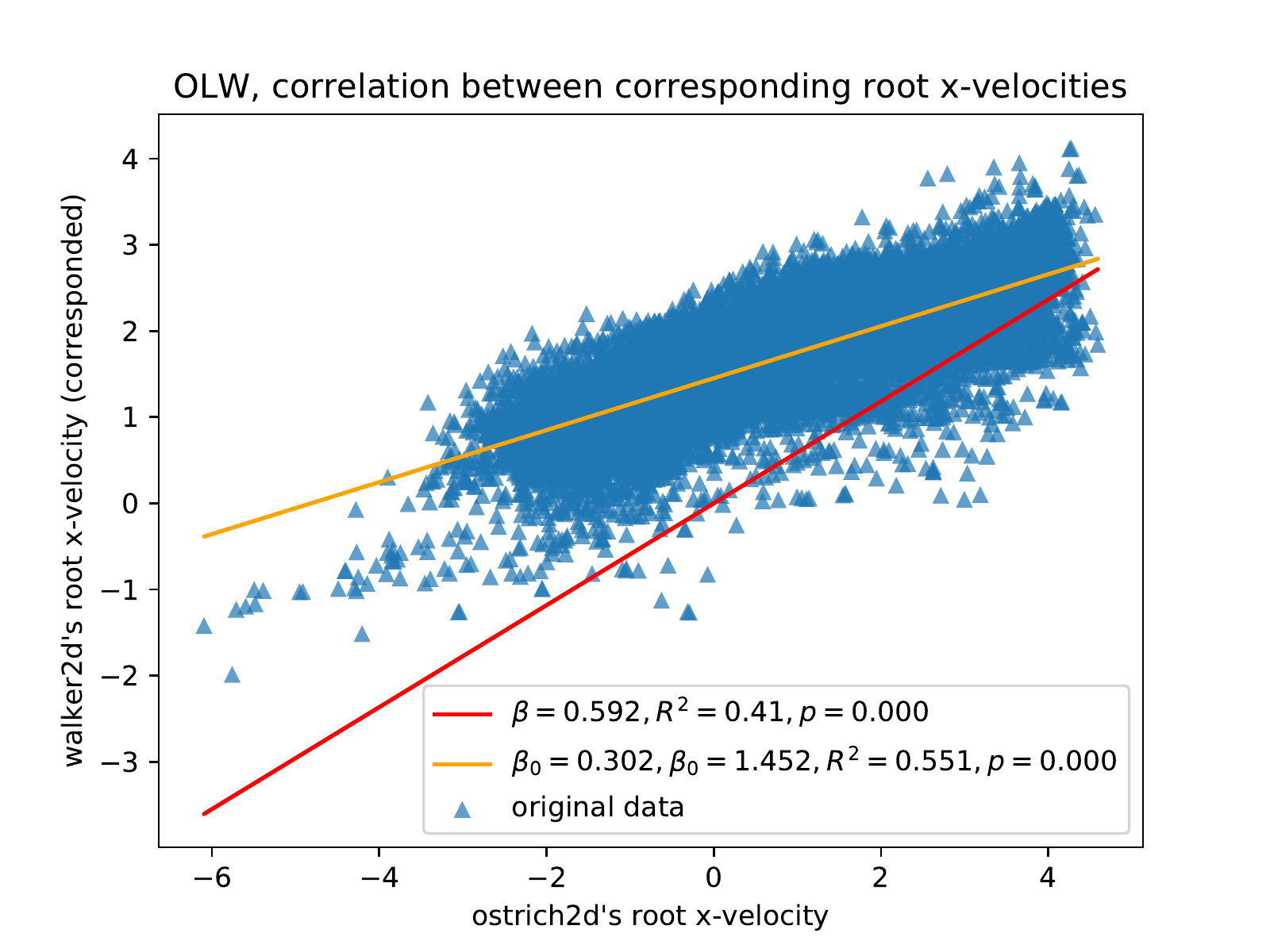}
\end{minipage}
\begin{minipage}[t]{.5\textwidth}
    \centering
    \includegraphics[width=\textwidth]{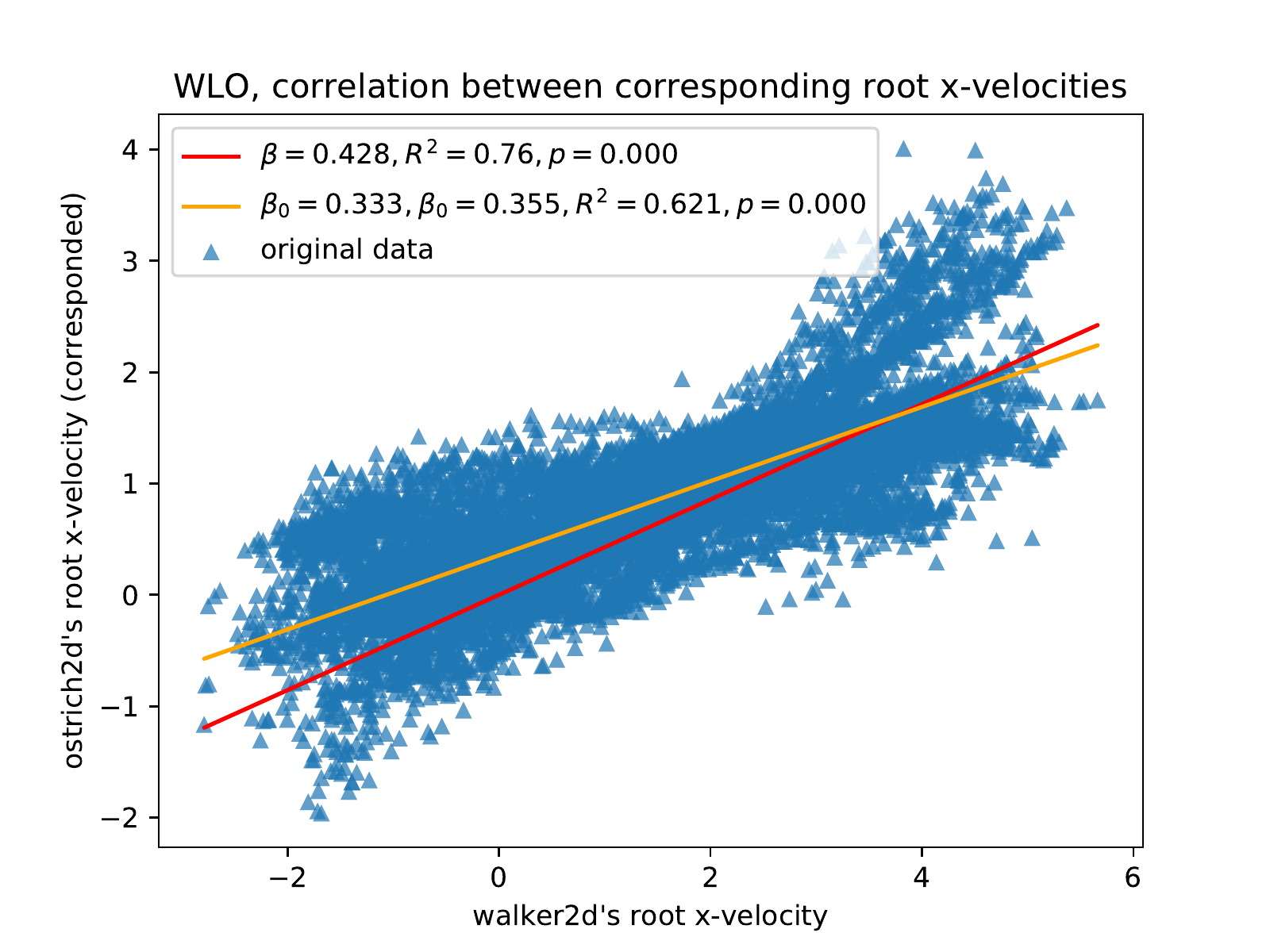}
\end{minipage}
\caption{Results of linear regression on the corresponded root x-velocity vs. the source system's root x-velocity. The lines represent the learned linear models. $\beta, \beta_0, \beta_1$ correspond to the parameters specified in Equations \ref{eq:lm} and \ref{eq:lm_intercept}. Red: without intercept. Orange: with intercept. Best viewed in color.}
\label{fig:corr}
\end{figure}
As mentioned, a true correspondence between dynamical systems must go beyond a simple mapping between two periodic trajectories. Examining perturbations is therefore crucial in assessing the validity of our method. We provide a brief statistical analysis to show that the perturbed states indeed map to each other in a meaningful way. To this end, we use the correlation between the original system's root x-velocity and the corresponded root x-velocity of the target system. We prepare simple linear models:
\begin{align}
    \hat{\dot x}^B_t &\sim \beta \dot x^A_t + \epsilon \label{eq:lm} \\
    \hat{\dot x}^B_t &\sim \beta_0 + \beta_1 \dot x^A_t + \epsilon \label{eq:lm_intercept}
\end{align}
where $\dot x^A_t \in s^A_t$ and $\hat{\dot x}^B_t \in \hat s^B_t$ are the root x-velocity of system A and the estimated corresponding system B state. We first sample $s^A_t$ from the state trajectory dataset. For each $s^A_t$, we generate 10 new data points by adding a uniform noise to $\dot x^A_t$ to create diverse scenarios where the root x-velocity is perturbed. We then predict $\hat{s}^B_t$ using L2CDS and finally learn the regression coefficients $\beta, \beta_0$, and $\beta_1$.

Figure \ref{fig:corr} summarizes the results of the above experiment on the $n=1000$ states gathered from \texttt{ostrich2d} and \texttt{walker2d}. In both OLW and WLO, the slope coefficient of the simple linear model is significantly above zero with the reported $p \ll 0.05$, whether the intercept is included or not. This shows that perturbations such as a forward or backward push on the original system results in a push in the same direction on the target system when L2CDS is used to estimate the corresponding target system states.

\section{Aperiodic Dynamical Systems}
\begin{figure}[!ht]
    \begin{minipage}[t]{.33\textwidth}
        \centering
        \includegraphics[width=\textwidth]{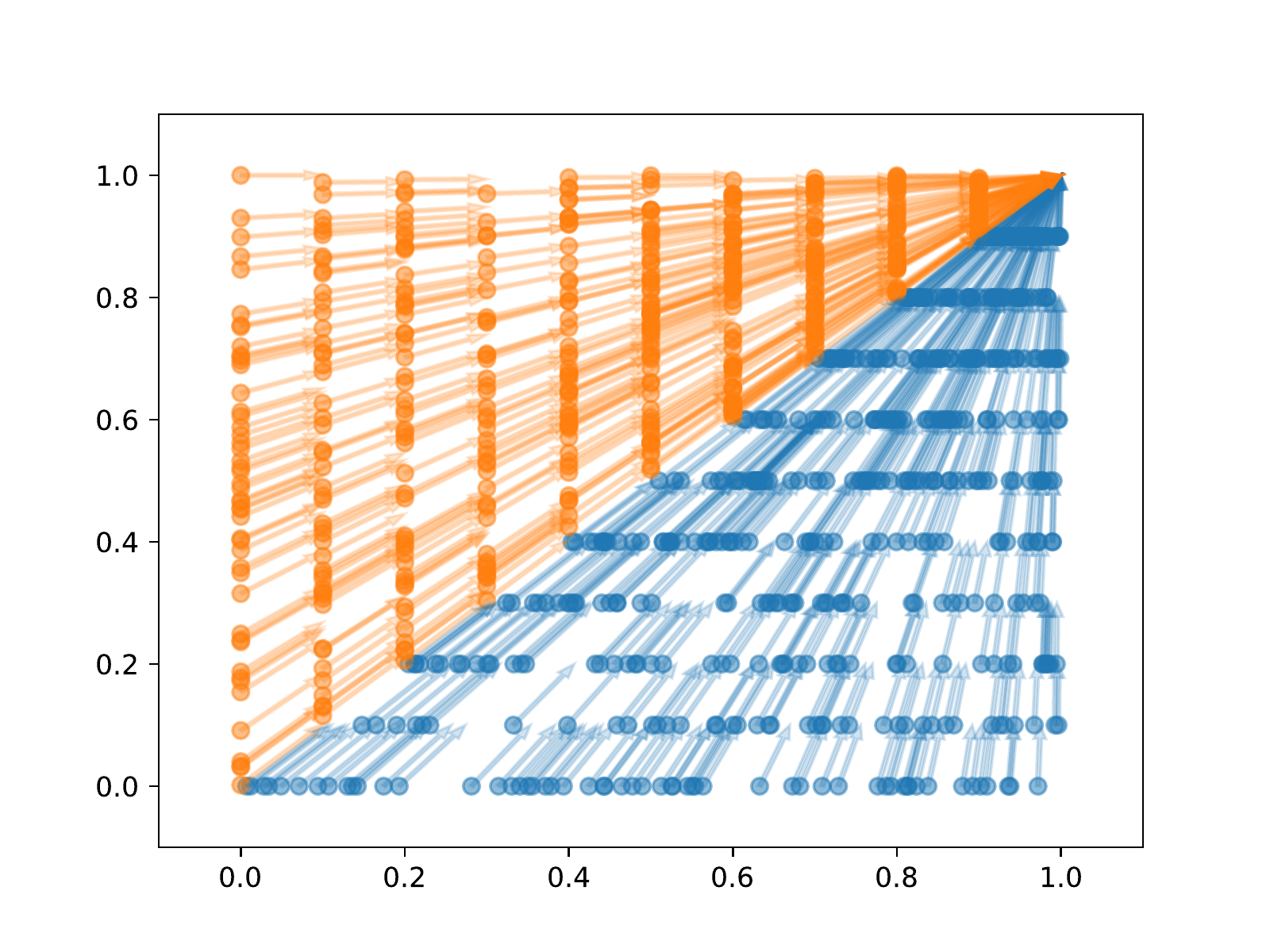}
    \end{minipage}
    \begin{minipage}[t]{.33\textwidth}
        \centering
        \includegraphics[width=\textwidth]{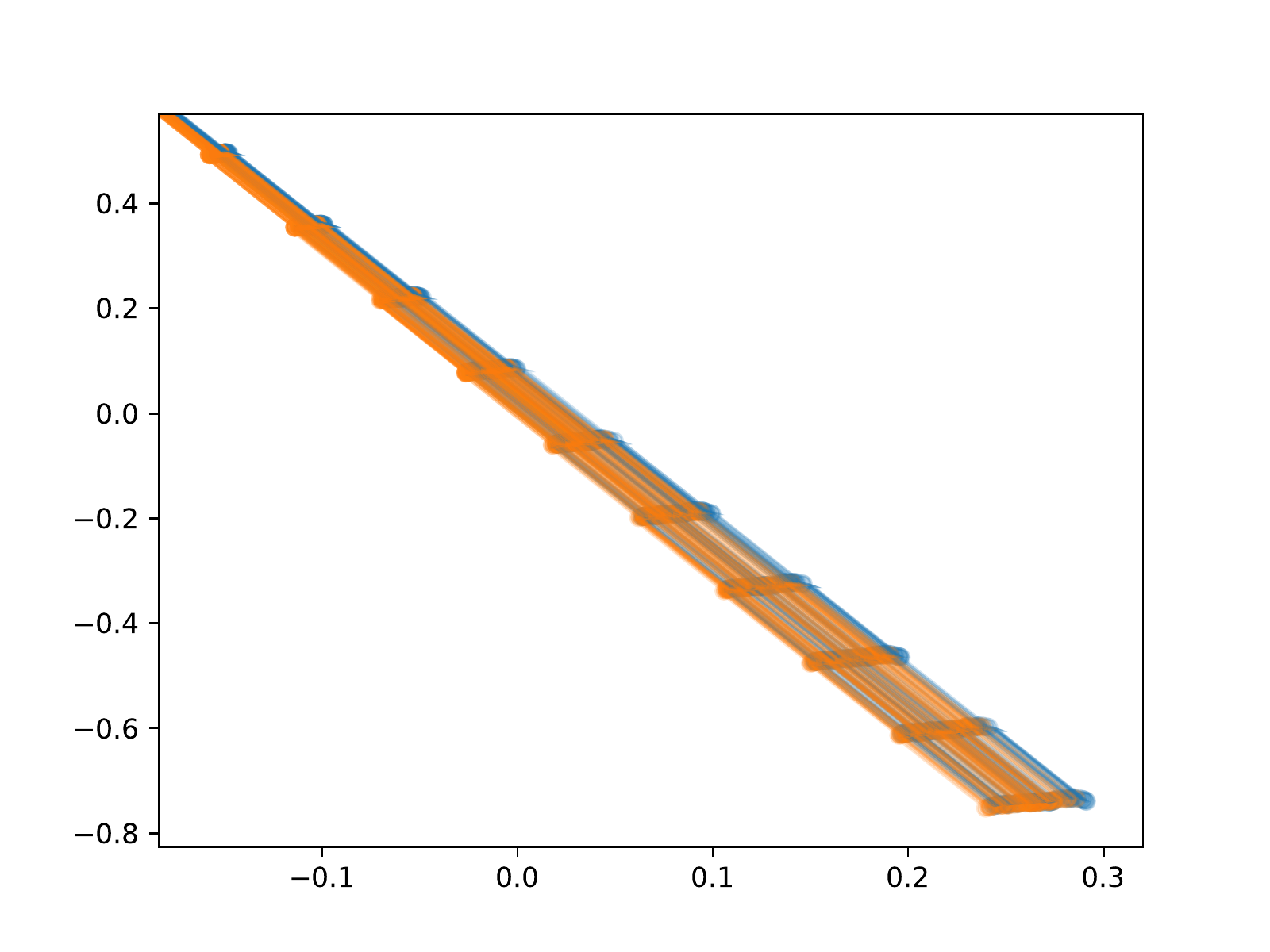}
    \end{minipage}
    \begin{minipage}[t]{.33\textwidth}
        \centering
        \includegraphics[width=\textwidth]{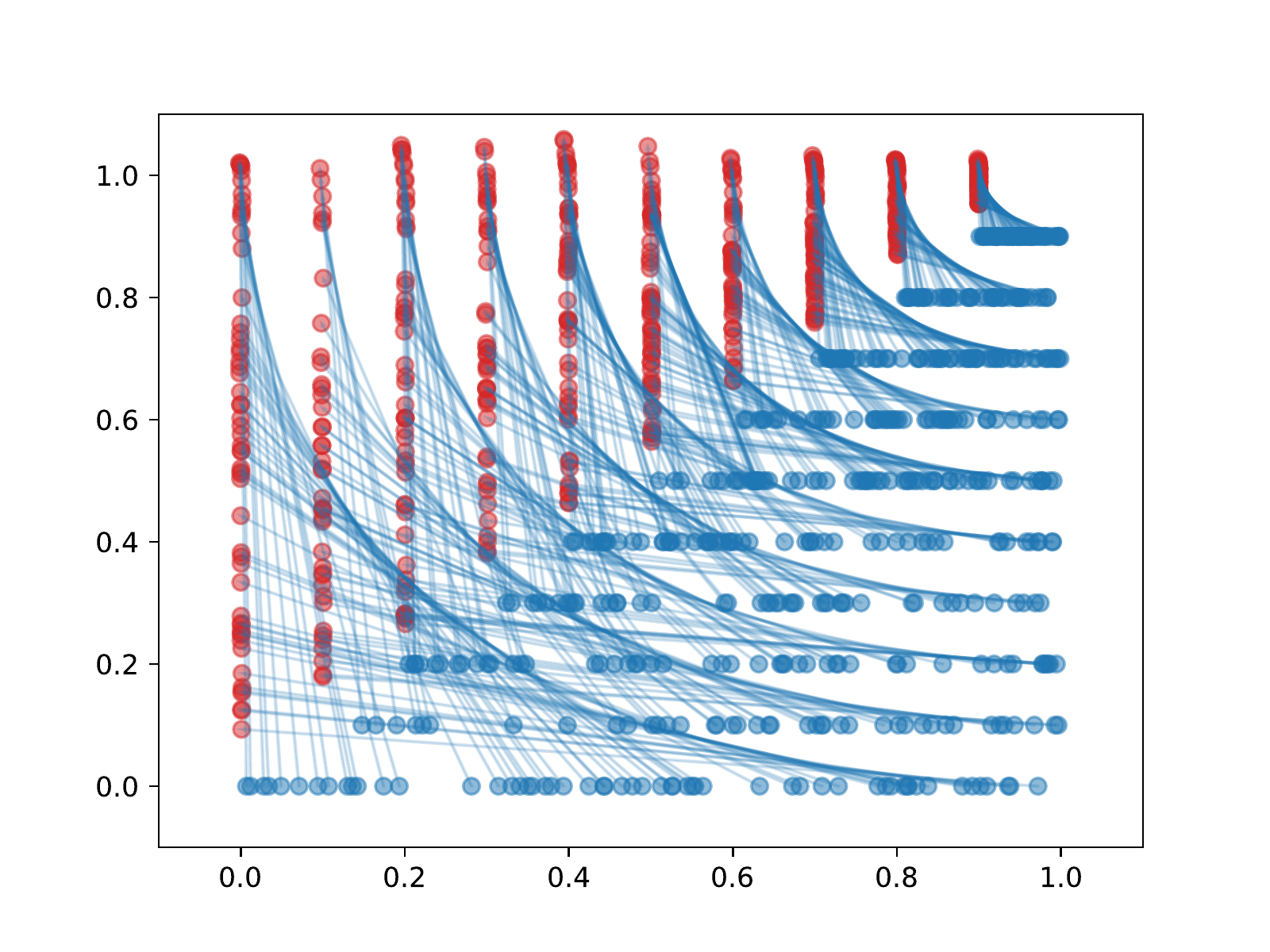}
    \end{minipage}
    \caption{Left: orange and blue points are two aperiodic dynamical systems, whose particles in motion towards a common point of convergence. Middle: the latent state resulting from applying L2CDS to the state tuples of the two systems. The output of the latent dynamics model is visualized as arrows. Right: visualizing the correspondence between the blue system and the orange system. The blue lines connect the original blue system states to the estimated corresponding orange system states (red points). Best viewed in color.}
    \label{fig:wedges}
\end{figure}

Our method can be applied to both periodic and aperiodic systems. Figure \ref{fig:wedges} illustrates two systems whose dynamics move particle towards a common point. The two systems truly correspond to each other as they are are mirror images. Using L2CDS, we retrieve this correspondence successfully. 

\section{Table of Hyperparameters}

\begin{table}[H]
    \centering
    \begin{tabular}{c|c|c|c|c|c|c}
     Experiment & $|s'_t|$ & $\lambda_{\mathrm{AE}}$ &  $\lambda_{\mathrm{NN}}$ & $\lambda_{\mathrm{FD}}$ & $\lambda_{\mathrm{PV}}$  & $\sigma$ \\
     \hline
     \texttt{walker2d} and \texttt{pendulum} & 2 & 1 & 1 & 1e-3 & 1e-3 & 1e-1 \\
     \texttt{ostrich2d} and \texttt{walker2d} & 8 & 1 & 1 & 1e3 & 1e3 & 5e-3 
    \end{tabular}
    \caption{For each experiment, we tune the following hyperparameters: $|s'_t|$: dimension of latent state. $\lambda_{\mathrm{AE}}, \lambda_{\mathrm{NN}}, \lambda_{\mathrm{FD}}, \lambda_{\mathrm{PV}}$: loss weight for autoencoder loss, symmetric nearest neighbors loss, latent forward dynamics consistency loss, and latent pose-velocity loss, respectively. $\sigma$: standard deviation for the Gaussian noise input for the decoders and the latent forward dynamics model.}
    \label{tab:hyperparameters}
\end{table}

\end{document}